# Physics-Informed Machine Learning for Microscale Drying of Plant-Based Foods: A Systematic Review of Computational Models and Experimental Insights

C. P. Batuwatta-Gamage[1]*, H. Jeong[1], HCP Karunasena[3], M. A. Karim[1], C.M. Rathnayaka[1,2], and Y.T. Gu[1]*

[1] School, Mechanical, Medical and Process Engineering, Faculty of Engineering, Queensland University of Technology (QUT), Australia.

[2] School of Science, Technology and Engineering, University of the Sunshine Coast (UniSC), Australia.

[3] Department of Mechanical and Manufacturing Engineering, Faculty of Engineering, University of Ruhuna, Sri Lanka.

* Corresponding authors from: Queensland University of Technology (QUT), Australia.

E-mail address: batuwatt@qut.edu.au (C.P. Batuwatta-Gamage) and yuantong.gu@qut.edu.au (Y. T. Gu)

## Abstract

This review examines the current state of research on microscale cellular changes during the drying of plant-based food materials (PBFM), with particular emphasis on computational modelling approaches. The review addresses the critical need for advanced computational methods in microscale investigations. We systematically analyse experimental studies in PBFM drying, highlighting their contributions and limitations in capturing cellular-level phenomena, including challenges in data acquisition and measurement accuracy under varying drying conditions. The evolution of computational models for microstructural investigations is thoroughly examined, from traditional numerical methods to contemporary state-of-the-art approaches, with specific focus on their ability to handle the complex, nonlinear properties of plant cellular materials. Special attention is given to the emergence of data-driven models and their limitations in predicting microscale cellular behaviour during PBFM drying, particularly addressing challenges in dataset acquisition and model generalization. The review provides an in-depth analysis of Physics-Informed Machine Learning (PIML) frameworks, examining their theoretical foundations, current applications in related fields, and unique advantages in combining physical principles with neural network architectures. Through this comprehensive assessment, we identify critical gaps in existing methodologies, evaluate the trade-offs between different modelling approaches, and provide insights into future research directions for improving our understanding of cellular-level transformations during PBFM drying processes. The review concludes with recommendations for integrating experimental and computational approaches to advance the field of food preservation technology.

***Key Words:*** Food drying; Heat Transfer; Mass transfer; Shrinkage; Physics-informed machine Learning

# 1 Introduction

Food insecurity and waste represent critical global challenges that demand urgent attention. According to the World Health Organization (WHO), these issues significantly impact human nutrition and well-being [1]. Recent statistics indicate that approximately 3.1 billion people lack access to healthy diets, while paradoxically, substantial amounts of food continue to be wasted globally [2]. Food serves as humanity's primary source of essential calories, nutrients, vitamins, and minerals, with plant-based foods being particularly crucial for providing vital micronutrients necessary for bodily functions, growth, and reproduction. The reduction of food waste has emerged as a pivotal strategy to address multiple challenges simultaneously. Beyond alleviating hunger, minimizing food waste contributes to reducing greenhouse gas emissions, conserving energy, and decreasing pressure on natural resources [3]. Studies estimate that food waste accounts for 8-10% of global greenhouse gas emissions, highlighting the environmental significance of this issue. In this context, food preservation techniques have gained considerable attention as effective solutions to extend food availability and reduce waste, particularly in regions with limited access to cold storage facilities.

Among various preservation methods, food drying stands out as one of the oldest and most widely adopted techniques, with roots tracing back to ancient civilizations [4]. This method operates on a fundamental principle: reducing moisture content in food materials to inhibit microbial growth and slow enzymatic reactions that lead to spoilage. The benefits of food drying extend beyond preservation, including reduced weight for efficient storage and transportation, as well as concentrated flavours in the final product. Furthermore, dried foods often retain a significant portion of their nutritional value when processed under optimal conditions. The evolution of drying technology has seen significant advancement from traditional methods to modern techniques [5]. While traditional sun and air-drying methods remain relevant in certain contexts, particularly in developing regions, contemporary approaches such as oven drying, spray drying, microwave drying, and freeze-drying offer enhanced control over critical parameters like temperature, humidity, and air circulation. These advanced methods enable producers to achieve more consistent quality and predictable shelf life in dried products. However, the selection of an appropriate drying method requires careful consideration of multiple factors, including the specific characteristics of the food material, desired quality outcomes, and available resources [6].

Despite these technological advances, understanding the fundamental mechanisms of food drying at the cellular level remains a significant challenge [7]. The complex nature of plant-based food materials, characterized by heterogeneous structures and varying composition, makes it difficult to predict and optimize drying processes accurately. This complexity is particularly evident in the microscale changes that occur during drying, which directly influence the final product's quality, nutritional content, and organoleptic properties. Recent developments in computational modelling and simulation techniques have opened new avenues for investigating these microscale phenomena, potentially leading to more efficient and effective drying processes.

The growing emphasis on sustainable food processing and the increasing demand for high-quality dried products has sparked renewed interest in optimizing drying technologies [4]. This optimization requires a deep understanding of both the macroscopic and microscopic aspects of the drying process, as well as the development of innovative approaches to address current limitations in food preservation techniques. Additionally, the integration of smart technologies and automation in food drying processes presents opportunities for improving energy efficiency and product quality while reducing operational costs [8].

## 2 Microstructural Characteristics and Changes in Plant-Based Food Materials During Drying

Experimental findings show that plant-based food materials (PBFM) contain substantial amounts of water, typically 80-90% of their mass, distributed throughout their hygroscopic microstructure. In fresh PBFM, most of this moisture (80-92%) is located within the cells (intracellular space). The remaining water is distributed between intercellular spaces (6-16%) and cell wall compartments (1-6%) [9]. As drying progresses, this water migrates to the PBFM's surface via various pathways, subsequently evaporating into the drying environment. Literature indicates two main water transport mechanisms. First, water within cells can move to intercellular spaces either through fine capillaries (apoplastic transport) or through cell rupture when drying temperatures exceed 50°C. Second, water can travel between adjacent cells through microcapillaries in the cell walls (symplastic transport). Subsequently, this water, along with moisture already present in intercellular spaces, diffuses to the surface where it evaporates into the drying medium [7]. The structural stability of fresh food microstructures depends on turgor pressure - internal cell pressure created by water content. As water is removed during drying, the loss of turgor pressure destabilizes these microstructures. This destabilization leads to uneven volume reductions (anisotropic shrinkage) and changes in porosity and density. These morphological changes, which begin at the cellular level, eventually manifest as bulk-level variations. Research has shown that these bulk-level morphological changes are significantly influenced by microscale cellular alterations, highlighting the importance of understanding microscale PBFM behaviour during drying. Since cells contain the majority of water, most experimental studies and computational models focus on single-cell changes during drying [10-15].

Researchers have employed various advanced techniques to study microscale variations in PBFM during drying [16]. These include scanning electron microscopy (SEM), x-ray microtomography, nuclear magnetic resonance (NMR), near infrared (NIR) spectroscopy, light microscopy, nanoindentation, and atomic force microscopy (AFM). These studies have examined water distribution within PBFM microstructures [17, 18] and investigated moisture transport mechanisms during drying [19, 20]. Studies using microphotographs and SEM images have documented 2D morphological changes in plant cells during drying, correlating these changes with moisture content [13, 15]. Advanced 3D imaging techniques have revealed morphological and surface roughness variations in plant cells as moisture content changes [14]. Additionally, nanoindentation experiments have measured micro-level mechanical properties such as stiffness, hardness, and elastic modulus throughout the drying process [21]. Recent research using X-ray micro-computed tomography has investigated microstructural changes including porosity, pore diameter, cell sphericity, cell elongation, and diameter variations through 3D imaging [22].

However, studying cellular-level variations in PBFM during drying presents several experimental challenges. The primary limitations include the need for sophisticated equipment and the time-intensive nature of these investigations. While some theoretical experimental studies exist, real-time investigations and measurements during drying often prove impractical. For example, NMR, while effective at tracking cellular-level water content, can suffer from data noise and limited sensitivity [16]. The difficulty of imaging during drying has led researchers to use separate samples for measuring variations at different moisture levels [13]. Although some studies have successfully documented changes in plant tissues during drying, they either rely on limited data from a single specimen [14] or require separate experiments for different measurements, such as 3D imaging for morphological changes and data logging systems for mass and temperature measurements [22]. These limitations make it impractical to rely solely on experimental methods for capturing microscale changes during continuous drying processes. Consequently, researchers have turned to physics-driven mathematical models for more comprehensive analysis of food microstructures during drying, as discussed in Section 3.

## 3  Physics-Based Models for Microstructural Investigations

Physics-based models, used to investigate changes within PBFM during drying, are typically formulated from foundational principles such as thermodynamic laws and mass and heat transfer equations [23]. The following subsections explain these fundamental principles in detail.

### 3.1  Equations based on mass conservation and continuity equations

Fick's Law is commonly used to describe the diffusion of a substance through a medium and is deeply rooted in the principle of mass conservation. The most basic form of Fick's first law for one-dimensional diffusion is expressed as [24]:

$$J = -D \frac{\partial c}{\partial x}, \tag{1}$$

Where, $J$ is the diffusion flux, representing the amount of substance that will flow through a unit area per unit time. $D$ is the diffusion coefficient, which depends on the medium and the substance. $\partial c / \partial x$ is the concentration gradient in the direction of diffusion.

In order to integrate Fick's first law with mass conservation, it can be considered the rate of change of concentration. $c$ in a given region must be equal to the net mass flux of particles into that region. For a three-dimensional system, this concept is mathematically described by the continuity equation for mass conservation, combined with Fick's First Law, resulting in Fick's Second Law [24]:

$$\frac{\partial c}{\partial t} + \nabla(-D\nabla c + uc) = S, \tag{2}$$

where, $u$ is convective flow velocity, and $S$ is the rate of moisture production or consumption.

This partial differential equation (PDE) describes how the concentration of a diffusing substance varies in time and space. In the context of food drying, Fick's Second Law can be adapted to account for the movement of moisture (or other solvents) out of a food material into the drying medium. Here, $c$ could represent the moisture concentration at various locations within the food, and $D$ would be the effective diffusivity of moisture within that specific food material. Thus, using Fick's Law based on the principle of mass conservation allows to model the diffusion process realistically, making it a valuable tool for understanding and optimizing food drying processes [25]. Fick's diffusion law can be coupled with heat transfer equations for more comprehensive models in food drying.

### 3.2 Equations based on energy conservation

In food drying processes, the primary mechanism for moisture removal often involves heat transfer. Just as Fick's Laws are rooted in mass conservation principles, the heat transfer equations governing food drying stem from energy conservation. The equation commonly used to describe heat conduction is Fourier's Law, which can be written as [26]:

$$\rho C_p \frac{\partial T}{\partial t} + \rho C_p u . \nabla T = \nabla (k \nabla T) + Q_e, \tag{3}$$

where $\rho$ is density, $C_p$ is specific heat, $T$ is temperature, $k$ is thermal conductivity, and $Q_e$ is the internal rate of heat generation.

### 3.3 Equations based on fluid mechanics

Fluid mechanics plays a significant role in food drying processes, especially in methods involving convective drying. By approximating the cellular fluid as an incompressible, homogeneous fluid, the Navier-Stokes (NS) equations can be applied to investigate fluid flow characteristics at both the cellular and microstructural levels. In state-of-the-art meshfree methods, field properties such as density, velocity, and acceleration of particles within the fluid domain are also calculated based on the NS equations [11]. Despite their potential applicability, fluid mechanics models have been less commonly employed than heat and mass transfer models in food drying analyses.

### 3.4 Equations based on solid mechanics

Solid mechanics provides an essential framework for understanding the structural changes in food products during drying, particularly the phenomenon of shrinkage. The field incorporates various equations to describe how materials deform under different loads, including stress-strain relationships and constitutive equations that specify material behaviour under conditions like elasticity, plasticity, hyper elasticity, and viscoelasticity. In the context of food drying, these equations are adapted to capture the unique, often complex behaviours of food items as they lose moisture [27].

One of the critical aspects of using solid mechanics in food drying is the coupling of these mechanical equations with those of heat and mass transfer. This integrated approach allows for a comprehensive understanding of the drying process, as it accounts not only for moisture and temperature changes but also for the physical and mechanical deformations that occur in the material. For example, as food dries, it experiences volume changes and internal stress development. Accurately predicting these requires an understanding of material properties like Young's modulus, Poisson's ratio, and shear modulus, which are all concepts derived from solid mechanics [3, 28, 29].

Analytical solutions for physics-based mathematical models are possible for one-dimensional (1D) or similar problems with simple geometries and physical properties. However, drying plant-based foods involves simultaneous and transient heat, mass, and momentum transfer in irregular and dynamically changing microstructures. This complexity necessitates multiphysics models that integrate coupled phenomena and nonlinear variations in material properties, along with computational models to accurately solve these multiphysics problems. The advancement of physics-based models using computational approaches for PBFM's complex microscale cellular analysis during drying is discussed in Section 4 onwards.

### 4 Computational modelling frameworks for microstructural investigations

PBFM often display irregular microstructures characterized by their hygroscopic, heterogeneous, and porous internal configurations. During drying, several parameters, including material properties, moisture content, and temperature gradients, change, often in a nonlinear manner. These complexities and nonlinearities make it challenging to derive analytical solutions from physics-based models and accurately capture the material's complex variations. Numerical computational methodologies are particularly effective at addressing these intricate nonlinear equations, providing a flexible toolkit to analyse both nonlinear dynamics and a wide range of conditions and variables. Consequently, computational approaches are routinely used to analyse these variances and optimize the drying process [27, 30, 31]. Sections 4.1.1 and 4.1.2 investigate microscale cellular variations in PBFM during drying, exploring the intricacies of mesh-based and meshfree models and highlighting their limitations.

### 4.1.1 Mesh-based computational modelling frameworks

The Finite Element Method (FEM) is a widely used numerical technique where the problem domain is discretized into smaller subdivisions, known as a mesh. These mesh elements are permanently interconnected through nodes. At each node, the governing partial differential equations (PDEs) are solved to approximate the solution, considering both the initial conditions (IC) and boundary conditions (BC) [32].

Over recent decades, models based on FEM have been effectively harnessed to emulate intricate mathematical frameworks and delineate microscale morphological variations in PBFM. As one of the first approaches for plant cell morphological investigations with FEM, cell wall material property variations of a simple circular shaped 2-D cell structure have been extracted identifying the shape variations along with turgor pressure by Smith, et al. [33]. In the model, cell wall is considered as permeable allowing volume loss during compression. Wu and Pitts [34] formulated a three-dimensional (3-D) FEM model for an apple parenchyma cell using actual cell geometry and thin shell finite elements for the cell wall, incorporating turgor pressure as a distributed force. The simulation outcomes aligned well with experimental data and existing analytical models, establishing its validity as an instrumental tool in studying apple cell attributes. Dintwa, et al. [35] developed a FEM model to simulate the compression of single tomato cells, accounting for features like permeable cell walls and internal fluid dynamics. Authors claimed that the model successfully replicated the force-deformation behaviours and shape changes of single cells under compression and may serve as a foundational element for broader tissue deformation simulations.

Complementing FEM in tissue studies, numerical analyses often incorporate virtual tissue microstructures. Mebatsion, et al. [36] initiated the concept of virtual microstructure development by proposing a microscopic image-based method for microstructure evolution. In particular, an ellipse tessellation algorithm is used integrating spatial and geometrical properties derived from microscopic images to devise a virtual 2-D microstructure for apple fruit. Abera, et al. [37] followed the concept to elucidate cellular architecture variations in pome fruits, portraying cells as closed structures influenced by turgor pressure with adjacent cell walls adhering to Hooke's law. Using Voronoi tessellation for initial topology and differential equations resolved via the ODE45 method for determining cell shape evolution, the model successfully creates realistic 2D fruit tissue structures for simulations pertinent to heat, mass transfer, or mechanical changes during controlled storage of fresh pome fruits.

This methodology has since become a staple in many FEM-based microscale studies on plant tissue drying as well as dehydration investigations [38-40]. Particularly, varying sizes and shapes of intercellular and intracellular spaces have been adopted to depict the microstructural heterogeneity in such 2-D representative microstructures. Aregawi, et al. [41] introduced a one-dimensional FEM model, coupling microscale deformations with a water transport framework tailored for apple tissue, integrating both linear elastic and nonlinear viscoelastic properties. These authors subsequently broadened their work into a 2-D axisymmetric model, aiming to gauge tissue shrinkage during apple tissue dehydration, emphasizing viscoelastic properties [42]. Fanta, et al. [40] utilised a cell growth technique to craft a water transport model for pear tissue, leveraging authentic tissue geometry and a prior-validated cell growth algorithm, marking a distinction from earlier plant tissue research. Their model underscored the pivotal role of microstructural features in water transport, highlighting the cell membrane's dominant influence on macroscopic water conductivity. These same authors then coupled a deformation model, offering a richer insight into the mechanisms of water loss. Specifically, they proposed an intricate 2-D microstructural deformation method, embedding elastic attributes into the mechanical framework. Notwithstanding its merits, the model presented a major limitation: it was only capable of simulating up to 30% water removal, falling short for food drying processes that typically require more than 50% water extraction.

Recently, Prawiranto, et al. [43] adopted a different modelling approach with FEM, incorporating a 3-D representative cellular models to investigate dehydration kinetics of apple cells, considering structural changes from turgid states to phenomena like shrinkage and membrane rupture. The model revealed that plasmolysis can decrease equilibrium water content by up to 60%, while lysis can quintuple tissue water permeability compared to other states. Notably, dehydrated tissue impedes the moisture removal rate, acting as a barrier against extracting moisture from the fresh tissue beneath. The study was later extended with a 3-D

Representative Elementary Volume (REV) to as an upscaling method, to elucidate the link between drying kinetics and the subsequent microstructural alterations in apple fruit during convective drying. The microscale analysis indicated that inducing lysis can elevate tissue permeability by up to four times, compared to mere cell shrinkage. When scaled up, the research pinpointed the formation of a low-permeability barrier layer during drying and determined that provoking lysis can enhance the drying rate more significantly (up to 26%) than drastic changes in drying conditions, underscoring the importance of considering cellular dehydration mechanisms in fruit drying studies [29]. Diverging from conventional FEM models, this study posits that it can effectively conduct analyses even under conditions of extremely low water activity levels.

Rahman, et al. [44] developed a microscale drying model that utilised a heterogeneous microstructure of PBFM to forecast the water transport mechanism at the cellular level during drying. Their approach harnessed SEM images to establish this heterogeneous microstructure, meticulously defining the cell walls, cells, and intercellular spaces as distinct vertices. Furthermore, authors provided a comprehensive explanation of the temperature profile within the food microstructure throughout the drying process, grounded in fundamental physics principles. By integrating a heat transfer model rooted in energy conservation with the mass transfer model, the micro-level temperature distribution in both cells and intercellular spaces was predicted.

After a thorough review of microscale analyses focused on PBFM's cellular variations and their integration with mesh-based computational methods, a methodical progression is evident. However, mesh-based FEM techniques come with inherent limitations due to their grid-centric domains and solver-dependent computations. A majority of these model simulations tend to rely excessively on oversimplified assumptions, notably the consistency and homogeneity of material properties and boundary conditions. Consequently, these models often fall short in representing critical cellular attributes, particularly non-linear deformations observed at extremely low moisture content levels or at intensive drying conditions. Furthermore, their ability to handle multiphase interactions is limited. The quest for a comprehensive drying model, one that can simultaneously address heat and mass transfer alongside shrinkage effects, continues. While mesh-based computational models do provide some valuable insights, they aren't fully equipped to deliver an exhaustive understanding of the transformations in plant cells or their microstructures during drying or dehydration [30]. Acknowledging such constraints, meshfree modelling techniques has been identified a notable substitute. Without the constraints of a grid-defined domain and enduring particle interactions, these methods are inherently more adaptable, adeptly addressing pronounced deflections and fluid shifts. The unique capabilities of meshfree models in this realm have been particularly highlighted in microscale drying research, as delineated in Section 4.1.2.

### 4.1.2 Meshfree particle-based computational modelling frameworks

In meshfree particle-based approaches, the problem domain is not discretised as it is in mesh-based FEM approaches. Instead, the domain is populated with a cloud of points or particles, which are generally distributed homogeneously across the domain. These particles represent both the geometry and properties of the system under investigation. Unlike the permanent nodal connections observed in mesh-based methods such as FEM, the interactions between particles in meshfree models are typically localised and temporary. Each particle interacts with its neighbouring particles based on specific criteria like distance or influence radius. These interactions can change dynamically during the simulation [45]. Such flexibility in local interactions is particularly advantageous for simulations involving large deformations in soft materials and fluid flow. The solving process in meshfree methods often involves constructing shape functions or influence functions around each particle to approximate the field variables, such as temperature, pressure, or displacement. These shape functions are then used in the formulation of governing equations, usually in integral form, to solve for the unknowns. Owing to the absence of a fixed, predefined mesh, meshfree methods are often more straightforward to implement in cases of complex geometries, multiphase interactions, and evolving problem domains with moving boundaries [46]. Researchers have exploited these capabilities to investigate microscale variations of PBFM during drying, thereby addressing challenges associated with mesh-based approaches [30, 47].

Karunasena, et al. [11] introduced meshfree modelling to analyse large shrinkage effects of plant cells during drying. Authors constructed a network of wall particles, each influenced by forces that correspond to various properties such as wall stiffness, wall-fluid repulsion, wall-fluid attraction, bending, and contraction. Simultaneously, the cell fluid was modelled as a highly viscous, incompressible Newtonian fluid exhibiting low-Reynolds-number flow characteristics. The behaviour of this fluid was shaped by forces related to pressure, viscosity, wall-fluid repulsion, and wall-fluid attraction. The computational model couples the Discrete Element Method (DEM) for simulating wall particles and the Smoothed Particle

Hydrodynamics (SPH) for fluid particles within the cell. Combined with a Lagrangian formulation, these methods augment the simulation's accuracy and efficiency. Simulations were conducted in discrete steps across the moisture-content domain. They continued until the magnitude of the turgor pressure nearly matched the magnitude of the initially set osmotic potential, halting further moisture exchanges through the wall. This moisture-content-based criterion was implemented to sidestep the prohibitive computational costs associated with small timesteps over the entire drying duration, a significant limitation in meshfree models developed for the food drying context. Combining the results with image processing techniques enables a visual representation of cell shrinkage. While the model can illustrate the wrinkling effects on cell walls, it does not offer physics-based explanations for such occurrences. Notably, despite the authors' claim that this 2-D meshfree cell model aligns reasonably well with experimental morphological findings, the model struggles to capture the nonlinear trajectory of morphological variations observed in experimental results. This underscores the challenges of predicting such intricate details using meshfree models. In an extension of their research, the authors introduced cell wall–fluid attraction forces and cell wall bending stiffness to address the notable shrinkage observed in plant cells during drying [10]. This addition of wall–fluid attraction forces and bending stiffness significantly improved the model's predictive capabilities. Expanding the study, the authors analysed various plant-based tissues, including apple, potato, carrot, and grapes. Two separate studies were undertaken to examine tissue shrinkage effects based on the initial cell shape, whether circular or hexagonal [48, 49]. The findings showed that cellular deformations are acutely influenced by factors such as cell dimensions, the physical and mechanical properties of the cell wall, properties of the middle lamella, and turgor pressure. The authors emphasised the model's proficiency in simulating tissues that experience substantial moisture content reductions, with the ability to achieve normalized moisture content values as low as 0.3. Further investigations provided insights into porosity development [50] and the effects of case hardening on tissue shrinkage and drying kinetics [51] of PBFM during the drying process.

Rathnayaka, et al. [12] subsequently introduced a SPH and Coarse-Grained (CG) numerical method to anticipate the deformation of 3-D cells in various PBFM. This research provided invaluable insights into the morphological behaviours of 3-D cellular structures, edging closer to a realistic representation. While the study contended that the model's predictions were generally in sync with experimental findings, it faced challenges in capturing the nonlinear behaviour observed in the morphological variations of those experiments, marking a significant shortcoming of this 3-D meshfree cell model. These investigations were later expanded to explore cell aggregations, aligning more closely with the actual experimental results of cellular structures [52]. The same researchers further refined the modelling framework to delve into the stress-strain-time relationships in plant cells, spanning from their fresh state to extreme dried conditions [53]. Later, Wijerathne, et al. [54] introduced a coarse-grained multiscale numerical model aimed at predicting bulk level (macroscale) deformations in PBFM tissues during the drying process. The study particularly highlighted that this method offers more accurate depictions of deformation behaviours and significantly reduces computational time, a notable drawback in earlier meshfree models. However, this model was found lacking in its ability to capture variations at the cellular level.

Meshfree models, as applied to microscale cellular investigations of PBFM during drying, have heralded several advancements. Primarily, these models excel at representing extensive deformations down to low moisture contents (i.e., up to 0.1 normalised moisture content), with an enhanced capacity for fluid-solid interactions. They also adeptly illustrate phenomena like cell wall bending, tightening, hardening, stretching, and interactions, allowing for visual depictions of complicated cellular morphological changes such as cell wall wrinkling. Moreover, the emphasis on advanced intercellular interactions and the distinct cellular nature of the microstructure positions these models a notch above their mesh-based counterparts. Nevertheless, meshfree models do present challenges. A significant drawback is the considerable computational time they require. This is largely attributed to the necessity to estimate neighbouring particles at each time step due to the temporary nature of available neighbouring particles [55]. Consequently, food drying simulations employing meshfree models are often constrained to a discrete moisture content domain, preventing analysis of morphological or any related variations in the time domain. Furthermore, there's a generalized assumption of constant hypothesised turgor pressure that linearly varies with moisture content, an approach that doesn't mirror reality. Further, meshfree models have developed only to capture morphological variations in the food drying context and no model yet developed to address other food-drying attributes defined along with heat and mass transfer variations [28, 56, 57].

Overall, numerical modelling, using traditional computational methodologies, undoubtedly offers deep insights into microscale changes in PBFM during the drying process. These insights often surpass what is achievable through experimental investigations alone. However, inherent challenges limit the full potential of these methodologies. One salient challenge shared by both mesh-based and meshfree techniques is the intricacy of discretising the heterogeneous and porous microstructures in a manner that strikes a balance between accuracy and computational efficiency. Beyond the above-mentioned limitations separately in Sections 4.1.1 and 4.1.2, neither method has shown the capability to incorporate the time-domain or drying-related variations, nor the non-uniform spatial-domain variation of material properties during microscale simulations in the drying context for PBFM. For example, expecting cell wall properties to remain constant during drying and homogenous within a realistic microstructure is unrealistic. Additionally, the variations and inconsistent boundary conditions in real drying scenarios have not been effectively addressed by either approach. Given these challenges, there's a noticeable pivot towards data-driven modelling approaches. These models leverage past real-world data, which inherently captures the nuanced spatiotemporal variations and the non-homogenous conditions. Such a comprehensive dataset naturally offers the potential for more accurate and holistic predictions in applications [8, 58]. The integration of data-driven models in examining microscale variations and their current status is elaborated upon in Section 5.

## 5   Data-driven modelling frameworks

As a result of the fourth industrial revolution (Industry 4.0) and advances in technology, such as sensors, computing resources, and cloud computing platforms, massive datasets have become available. These advances have led to novel modelling techniques that leverage statistical, computational, mathematical, and engineering theories to predict future outcomes based on past experiences [59]. These data-driven models have the potential to surpass traditional computational modelling in complex real-world applications. They are advantageous because they incorporate all the dynamic variations of properties and other factors, which are often too complex for traditional human reasoning along with physics. Consequently, predictions from well-trained data-driven models tend to be more realistic for real-world problems compared to physics-based models that rely on assumptions [60]. Hence, machine learning based data-driven approaches have been widely used to investigate various aspects of PBFM during drying [8, 58].

Most data-driven models in existing literature focus on macroscale analyses or higher. Khan, et al. [61] claimed to have developed a machine learning model for assessing the micromechanical properties of PBFM during drying. However, the artificial neural network (ANN) in this study is trained with only 76 data points, presenting significant challenge of underfitting to the training data. The absence of data-driven models for microscale investigations in PBFM likely arises from the difficulty of obtaining sufficient, high-quality datasets for such analyses. Experimental limitations, such as the inability to capture microscale variations during the drying process, repeatability issues, and uncertainties, add further complexity. Even the available data can be noisy and ill-posed. Further, "black-box" nature of data-driven modelling fails to interpret predictions by complex physics in science and engineering applications [62]. Although the practical advantages of data-driven framework offer sufficient justification for its use, its credibility as a dependable modelling tool ultimately hinges on its consistent alignment with physical principles and the transparency of the model. These challenges make it difficult to satisfy the data requirements for machine learning-based models in microscale investigations during PBFM drying, a limitation that extends to many other fields in engineering and science [63, 64].

## 6   Physics Informed Machine Learning (PIML) frameworks

### 6.1   PIML modelling framework

As the challenges of obtaining sufficient and high-quality data for microscale investigations in plant-based food materials are encountered with pure data-driven approaches, PIML is being recognised as a promising avenue. In this emerging approach, the predictive capabilities of machine learning models are combined with the governing principles of physics [65]. The fundamental equations that govern physical behaviours, as well as relevant problem-based conditions, are incorporated into these models as illustrated in Figure 6-1.

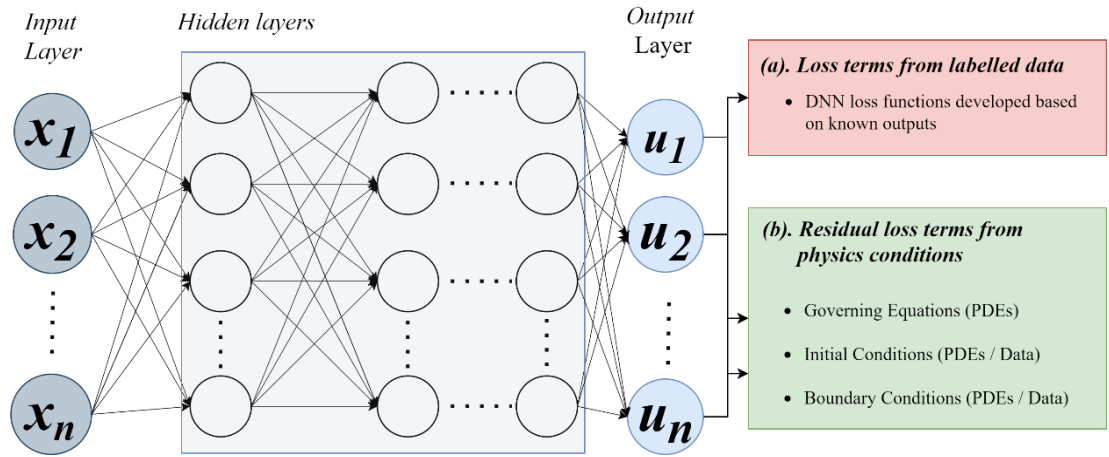

Figure 6-1: Physics-informed Machine Learning (PIML) that can couple data and physics through loss terms

PIML can be considered as a unique category of ANNs. As illustrated in Figure 6-1, the loss terms of the Neural Network (NN) can originate from two sources: labelled data and physics-based models. When the loss terms are solely based on labelled data (as in loss term (a) in Figure 6-1), the network functions as a typical data-driven model, such as a Deep Neural Network (DNN). However, when physics-based model loss terms are incorporated into the loss function, the approach is defined as PIML. This can further be divided into two main types: PIML with loss terms from both labelled data and physics, and PIML with loss terms only from physics-based models [57].

When PIML integrates loss terms from both labelled data and physics-based models, it effectively addresses the drawbacks often adhere with purely data-driven and traditional computational models separately. PIML excels in assimilating information from both labelled data and governing physics, which enhances predictability and reduces the data needed to attain a specific accuracy level [63]. Intriguingly, PIML can uncover previously unknown physical principles using existing real-world data. In this scenario, data compensates for the gaps in established physics, while the physics acts as a stabiliser for the model in ill-posed and noisy environments [66]. For instance, Yang, et al. [67] introduced a Bayesian physics-informed neural network to mitigate overfitting challenges that arise due to limited data and significant noise. He, et al. [68] illustrated the ability of PIMLs in estimating intricate parameters, especially in the context of sparse data. Pang, et al. [69] highlighted the efficacy of PIML in resolving multidimensional fractional advection-diffusion equations with notable precision, even when faced with noisy and sporadic data. Raissi, et al. [66], one of the pioneering researchers behind the PIML concept, demonstrated the model's remarkable capability in extracting velocity and pressure fields directly from visual data. This underscored the potential of PIML, in tandem with qualitative data and physics, to quantify physical parameters. Additional researches were showcased the synergistic benefits of combining physics with pertinent data in areas like heat transfer analyses [70-72], mass transfer [73], solid mechanics [74, 75], fluid mechanics [66, 76, 77], and other domains [63, 78-80].

Conversely, when the neural network's loss terms are based solely on the residuals of Partial Differential Equations (PDEs), boundary conditions (BCs), and initial conditions (ICs) (i.e., no loss terms from labelled data), the network is trained to adhere strictly to integrated physics definitions [81, 82]. This approach is similar to solving PDEs with respect to ICs and BCs in traditional computational methods [83]. In particular, the differential operators in PDEs are determined through Automatic Differentiation (AD), a contrast to conventional numerical methods like Finite Element Analysis (FEA) and meshfree techniques that rely on numerical differentiation [84]. Moreover, PIMLs do not necessitate a mesh or the handling of particle interactions for derivative calculations. Any point within a defined problem domain can be independently used in the associated PIML, devoid of any grid or node interactivity. Once trained for a specific spatiotemporal domain, a PIML can produce interpolations, extrapolations, or varying domain resolutions without additional training [85]. Hence, PIMLs offer particular advantages in solving PDEs compared to traditional computational methods, despite not being the most computationally efficient [86]. Blechschmidt and Ernst [87] have explored such supplementary capabilities by solving PDEs modelled with high-dimensional domains which have not yet been analysed with traditional computational methods. The capabilities of PIML computational approach has been extensively demonstrated and discussed in various fields and applications, including solid mechanics [65, 88-90], fluid mechanics [91, 92], heat transfer [93-95], and various other applications [96, 97]. The potential of PIML-based AD in solving PDEs has been probed across diverse scientific and engineering fields, adding unique value as a machine learning-based computational approach.

## 6.2  PIML modelling framework development directions

The development of PIML models has been greatly facilitated by the advent of robust machine learning platforms and libraries. TensorFlow [98] and PyTorch [99] stand out as two of the most popular frameworks at the moment, offering extensive capabilities for designing and training neural networks, as well as functionalities that are amenable to incorporating physical equations into the training process. For instance, TensorFlow offers automatic differentiation, a crucial feature for implementing the physical constraints of PIML. Libraries such as SciPy and NumPy also play a role in numerical computations involved in solving the physical equations. Specialised libraries have been developed by researchers, such as NeuroDiffEqn [100], IDRLnet [101], PyDEns [102] in PyTorch backend and SciANN [103], TensoDiffEq [104], and Elvet [105] in TensoFlow backend. These libraries are explicitly developed for PIMLs, offering pre-built modules and functions that simplify the implementation of PIML models. In addition to open-source platforms, commercial software like MATLAB [106] has also begun to integrate PIML capabilities, although often at a higher cost. These platforms feature built-in tools for data preprocessing, model building, and post-processing of results, which streamlines the entire model development process. They are also compatible with high-performance computing resources, a particularly crucial feature given the computationally intensive nature of PIML models. Furthermore, many researchers have published their code on GitHub, a cloud-based service [107, 108]. These development platforms and comprehensive code publications have been instrumental in accelerating research and applications in the PIML domain.

## 6.3   Exploring the capabilities of PIML modelling framework for analysing plant-based cells during drying

As highlighted in Section 4, traditional computational methods, with their oversimplified assumptions, have fallen short in thoroughly analysing microscale variations in PBFM drying. The primary reasons are method-specific challenges and the absence of vital property data. These models often overlook the variations in physical properties of the microstructure, boundary conditions, and inconsistent microstructural geometrical features. These aspects are crucial, especially as microstructural properties dynamically change with the moisture content during the drying process [28].  The merit of using data-driven approaches lies in the reality of the data, holding all relevant properties and their dynamic variations. Yet, as detailed in Section 2.5, obtaining a sufficient dataset for microscale variations is difficult due to the complexities in methodology and equipment. Despite advanced setups, continuous microscale behaviours cannot be reliably captured during continuous drying processes. It's noteworthy that while variations in tissues during drying have been observed. This why different samples have been used by Karunasena, et al. [13] and Mayor, et al. [15] to experimentally study microscale cellular variations. While cellular level variations within the same samples tissues during drying and dehydration have been observed by Rathnayaka, et al. [14] and Prawiranto, et al.

[22], a comprehensive extraction of all necessary changes from those sample variations to sufficiently support a data-driven model has not been achieved. Consequently, given the technological constraints, extracting adequate data for purely data-driven models for interpretable predictions remains an uphill task.

The recently introduced PIML framework emerges as a possible avenue. It offers a novel computational technique, depart from both traditional mesh-based and meshfree computational methods, as well as purely data-driven prediction models. By integrating physics-based models with relevant data, PIML could be possible to overcome limitations of both computational as well as pure data-driven approaches discussed in Sections 4 and 5. Moreover, this coupling sheds light on previously unidentified physical phenomena while addressing ill-defined conditions as discussed in Section 6. PBFM' microscale studies during drying are suffered by doubts in physical parameters like turgor pressure, osmotic pressure, cell wall stiffness and their dynamic variations with moisture content. PIML can potentially fill these knowledge gaps. Furthermore, PIML stands out in its ability to address non-linear conditions and soft matters without permanent mesh and particles [63, 66]. The blend of PIML's computational technique with a NN backend brings the model with considerable flexibility. This allows PIML to strengthen computational capabilities with the innate capabilities of statistical and NN methods. Such adaptability can be invaluable for tackling complicated and heterogeneous microstructure related challenges without being bound to the solver limits as well as other restrictions of traditional computational strategies.

Since its introduction in 2019, PIML's potential has been showcased in various fields [62, 63, 79]. It holds significant promise for advancing research in the food drying sector, particularly in studying microscale variations, a domain that faces challenges such as limited experimental data, scarce property information, measurement limitations, and unresolved physical definitions. Yet, to the best of the author's knowledge, as of this thesis's drafting, no research has been published by any other authors exploring the application of PIML.

## 6.4   PIML modelling in food drying

The authors of this paper have started exploring PIML computational methods for food drying, offering an alternative to current computational models and exploring its inherent capabilities [28, 56, 57, 109, 110]. These initiatives aim to tackle existing challenges in the field.

Initially, a PIML surrogate model was created to study mass transfer and shrinkage in plant cells, showcasing the model's strong predictive abilities in dealing with unknown conditions and its capability to integrate physics-based equations with DL techniques [28]. A novel PIML surrogate framework was developed that integrates two complementary models: PINN-MC for predicting temporal moisture concentration using water transport physics, and PINN-S for predicting moisture-dependent shrinkage effects based on the free shrinkage hypothesis. This work represents the first implementation of PIML modelling in food engineering applications. The physics-informed training approach of PINN-MC significantly mitigated overfitting compared to conventional deep neural networks while successfully handling uncertainties in predicting behaviour within unknown domains. Although the model required substantial training iterations, PINN-MC delivered robust and accurate moisture concentration predictions that remained consistent across varying drying parameters and food microstructures. The PINN-S component demonstrated that incorporating measurable moisture-mass loss data enabled more physically realistic shrinkage predictions, leading to the identification of a free shrinkage proportionality constant specific to the studied dataset. The integration of moisture concentration and shrinkage predictions within a unified framework effectively addressed the limitations inherent to individual modelling approaches. The results indicate that the PIML framework, constrained by relevant physical principles, can provide accurate real-time predictions for food drying processes using minimal experimental data, even under unknown conditions and nonlinear process variations. This successful fusion of multiple physical phenomena through surrogate modelling represents a significant step toward establishing a benchmark coupling methodology for industrial food drying processes, demonstrating clear advantages over traditional approaches in combining observational data with fundamental physics.

Following this, a PIML model was developed to assess the effectiveness of Automatic Differentiation (AD) in solving a Fickian diffusion mass transfer model for single-cell domain without labeled data [57]. The developed PINN-MT framework incorporated loss weights and adaptive hyperparameters in the activation function, achieving remarkable accuracy with errors below 0.23% and 0.33% for mass and moisture concentration predictions, respectively. The implementation of transfer learning significantly enhanced computational efficiency by approximately 80%. The framework maintained robust performance when predicting moisture content across varying cell-wall mass transfer coefficients and diffusivities, with errors remaining below 2.10%. Furthermore, the centralized PINN-MT showed excellent interpolation capabilities with errors under 1.96%. This pioneering work, representing the first application of PIML AD to food drying mass transfer, establishes a computationally efficient approach for real-time predictions under varying process conditions, addressing limitations of conventional numerical methods.

This research was later extended to solve coupled mass transfer and shrinkage models within a single-cell domain simultaneously, presenting a PIML-based computational framework for modelling nonlinear and heterogeneous morphological changes in plant cells during drying [110]. The framework integrated two specialized models: PINN-MT for mass transfer and PINN-NS incorporating Neo-Hookean hyper-elastic properties for nonlinear shrinkage predictions. Under homogeneous conditions, the coupled models successfully captured the interplay between mass transfer and cellular deformation, accounting for dynamic area variations, internal pressure changes, and resultant forces. The framework was further enhanced through domain decomposition to accommodate non-uniform cell-wall properties and heterogeneous boundary conditions. This advanced approach accurately predicted nonlinear shrinkage behaviour, heterogeneous cell area variations, cell wall thickness reduction, and physics-guided wall wrinkling at extremely low moisture content levels ($X/X_0 = 0.053$). The framework's ability to model these complex phenomena, particularly through domain decomposition, represents a significant advancement over traditional mesh-based and meshfree computational methods. Beyond its immediate application to plant cell modelling, this PINN-based approach establishes a new paradigm for predicting nonlinear and heterogeneous property variations in soft matter systems.

## 6.5 Limitations and drawbacks of PIML modelling framework

While PIML offers significant advantages, it also has several limitations and challenges. One major hurdle is the need for a thorough understanding of the underlying physics to accurately integrate physical laws into the model. Currently, most research and development rely on basic Python programming, where precise implementation of physical equations, derivative computations, domain definitions, and boundary conditions is critical. Any inaccuracies or omissions in these equations can result in misleading or erroneous outcomes.

Additionally, while PIML is designed to generalize better from sparse data, its effectiveness can still be compromised if the data is excessively limited or inherently noisy [111]. Random initialization can also lead to variations in final predictions, though these differences are often negligible. Using weight initialization through transfer learning can reduce such inconsistencies [112]. Integrating physical equations across multiple dimensions into machine learning frameworks can further increase computational complexity, potentially diminishing the efficiency benefits that PIML aims to provide. However, techniques such as customizing loss weights [113] or employing adaptive loss weight strategies [114] can help address these challenges. Another concern is model interpretability. Although physics-based constraints can enhance the reliability of predictions, they may obscure which features the model is leveraging, thereby reducing transparency. Finally, as PIML is still in its early stages of development, much remains to be explored regarding its scalability, reliability, and applicability across diverse food processing applications [85, 86].

# 7 Conclusion

Understanding the underlying behaviours and changes in properties of PBFM during drying is critical for optimising energy and time efficiency. Especially in PBFM, where bulk-level changes are primarily governed by microstructural behaviours during drying, understanding the fundamental microscale variations is crucial. Computational modelling has been recognised as a valuable tool for analysing this complex drying kinetics and exploring microscale behaviours. However, existing work has not successfully captured the highly nonlinear and heterogeneous variations in food microstructures. Mesh-based FEA has been widely used to explore various aspects of microscale variations during drying, yet it struggles with grid-based limitations. Meshfree frameworks, touted as an alternative to FEA, are also inefficient due to computational time constraints arising from interactions between neighbouring particles. Although meshfree models have demonstrated potential in capturing large morphological variations, their investigations have been limited to discrete ranges within the moisture content domain and focused solely on morphological aspects. Overall, both mesh-based and meshfree, fall short in capturing complex, nonlinear microstructural behaviours due to solver-based limitations. These models often assume homogeneous conditions and uniform drying kinetics, diverging from real-world scenarios. Purely data-driven machine learning models also face challenges due to insufficient and noisy data for PBFM microscale investigations.

Recently introduced PIML offers a promising alternative. As discussed in Section 6, PIML successfully integrates physics-based models with data-driven approaches, leveraging automatic differentiation to solve highly nonlinear behaviours. Its flexibility and the inherent capabilities of neural networks make it particularly adept at solving multi-domain and high-dimensional problems. Overall, PIML's ability to couple observational data with relevant physics offers distinct advantages for investigating microscale variations in PBFM during drying, particularly when the physics are not fully defined, and data are scarce.